\documentclass[11pt,letterpaper]{yalearxiv}

\usepackage{graphicx}
\usepackage{float}
\usepackage{bbm}

\usepackage{microtype}

\usepackage{natbib}
\setcitestyle{square}

\usepackage{subcaption}
\usepackage{booktabs}

\usepackage{amsmath}
\usepackage{amssymb}
\usepackage{mathtools}
\usepackage{amsthm}
\usepackage{dsfont}
\usepackage{multicol}
\usepackage{makecell}
\usepackage{multirow} 
\usepackage{amsfonts} 
\usepackage{mathrsfs}
\usepackage{enumitem}
\usepackage{pgfplotstable}
\pgfplotsset{compat=1.18}
\usepackage{lipsum}		

\usepackage{microtype}
\usepackage{graphicx}
\usepackage{booktabs} 
\usepackage[table]{xcolor}
\usepackage{arydshln}
\usepackage[normalem]{ulem} 

\usepackage{cases}
\usepackage{wrapfig}

\usepackage{url}

\usepackage{thmtools}
\usepackage{thm-restate}
\usepackage{tabu}

\definecolor{huskypurple}{HTML}{4B2E83}

\usepackage{titletoc}

\usepackage{listings}
\lstdefinestyle{promptstyle}{
  basicstyle=\ttfamily\footnotesize,
  breaklines=true,
  breakautoindent=false,
  breakindent=0pt,
  postbreak=\mbox{\textcolor{gray}{$\hookrightarrow$}\space},
  columns=fullflexible,
  keepspaces=true,
  frame=single,
  framesep=5pt,
  xleftmargin=6pt,
  xrightmargin=6pt,
  aboveskip=8pt,
  belowskip=8pt,
  showstringspaces=false,
}   
\makeatletter
\def\munderbar#1{\underline{\sbox\tw@{$#1$}\dp\tw@\z@\box\tw@}}
\makeatother

\AddToHook{cmd/appendix/before}{%
  \setcounter{axiom}{0}%
}

\newcommand{\be}{\begin{equation}}
\newcommand{\ee}{\end{equation}}
\newcommand{\bea}{\begin{equation*}\begin{aligned}}
\newcommand{\eea}{\end{aligned}\end{equation*}}

\newtcolorbox{simpleElegantQuote}{
    colback=AliceBlue!50!White,
    colframe=RoyalBlue!75!Black,
    boxrule=0.5pt,
    arc=2mm,
    boxsep=4pt,
    left=10pt, right=10pt,
    top=8pt, bottom=8pt,
    fontupper=\itshape,
}

\title{PhyCheck: Fine-Grained Evidence-Grounded Dataset for Physical Law Understanding in Video-LLMs}

\usepackage{fontawesome5}   

\definecolor{yaleblue}{RGB}{0,58,112}

\newcommand{\corresponding}{*}
\author{
    Zhongjie Ba\textsuperscript{\rm 1,\rm 2}, 
    Shengwang Xu\textsuperscript{\rm 1,\rm 2}, 
    Peng Cheng\textsuperscript{\rm 1,\rm 2}\corresponding, 
    Jinyang Zou\textsuperscript{\rm 1,\rm 2}, 
    Ting Yu\textsuperscript{\rm 3},
    Zhibo Wang\textsuperscript{\rm 1,\rm 2}, 
    Zhan Qin\textsuperscript{\rm 1,\rm 2}\\
    \textsuperscript{\rm 1} State Key Lab. of Blockchain and Data Security, Zhejiang University, Hangzhou, China\\
    \textsuperscript{\rm 2} Hangzhou High-Tech Zone (Binjiang) Institute of Blockchain and Data Security Hangzhou, Zhejiang, China\\
    \textsuperscript{\rm 3}Mohamed Bin Zayed University of Artificial Intelligence, Abu Dhabi, United Arab Emirates \\ 
    zhongjieba@zju.edu.cn, xushen9wan9@gmail.com, \{peng\_cheng, zou.jinyang\}@zju.edu.cn, \\ ting.yu@mbzuai.ac.ae, \{zhibowang, qinzhan\}@zju.edu.cn \\
\faGithub~\textbf{Source Code:} \href{https://github.com/masunozomi/PhyCheck}{\texttt{https://github.com/masunozomi/PhyCheck}}
}

\hypersetup{colorlinks=true, linkcolor=blue!50!black, citecolor=blue!50!black,
            urlcolor=blue!50!black}

\begin{document}


\begin{abstract}
\vspace{3mm}
{\large\textbf{Abstract}}

Embodied intelligence and world models require video understanding systems to go beyond recognizing objects and actions and develop an understanding of physical regularities. However, despite their strong performance on general video understanding tasks, current video-language models still struggle to reliably determine whether an observed event conforms to specific physical laws. Existing benchmarks primarily assess the physical quality of generated videos, providing limited support for systematically evaluating and improving the physical-law understanding of Video Large Language Models (Video-LLMs). To address this gap, we introduce \textbf{PhyCheck}, a video question answering dataset organized at two complementary levels of granularity. The coarse-grained subset asks models to determine whether the phenomenon shown in a video conforms to or violates physical laws, while the fine-grained subset further examines whether models can capture physical details responsible for the violation or compliance. We use these subsets as structured supervision to improve physical understanding. In addition,
the dataset contains a diagnostic subset with external causal context that reveal hidden factors affecting physical plausibility, assessing whether models can recalibrate their judgments accordingly. Experiments with Fine-tune Qwen2.5-VL show that training with the proposed data substantially improves the understanding of physical-consistency, while evaluations in the diagnostic subset reveal that current models still have difficulty incorporating additional causal conditions into their decisions. These findings highlight the gap between recognizing surface-level inconsistencies and understanding underlying physical mechanisms, and provide a foundation for evaluating and improving physical understanding in Video-LLMs.
\end{abstract}

\maketitle

\section{Introduction}
\begin{figure}[ht]
\centering
\includegraphics[width=0.8\linewidth
, height = 8cm
]{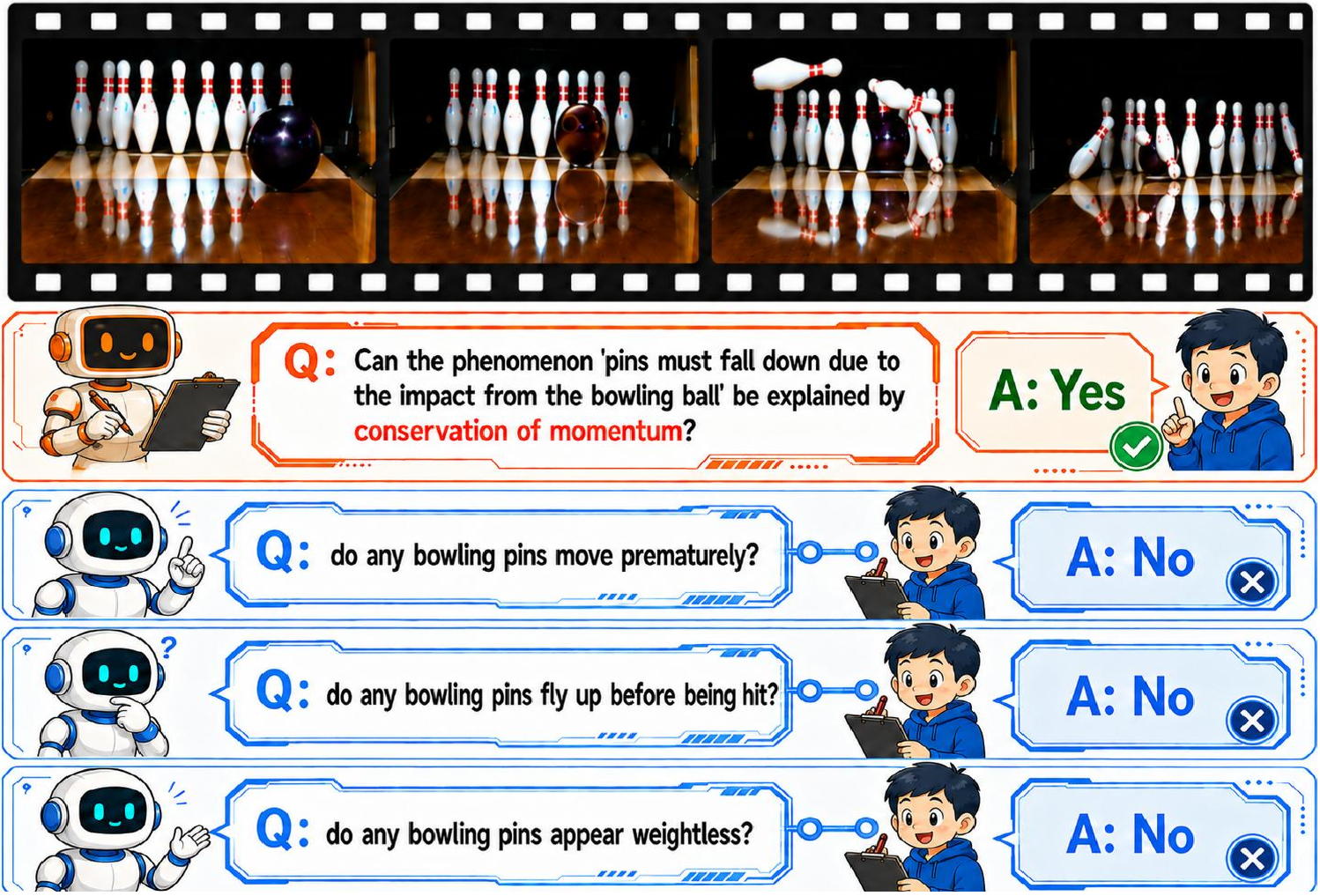}
\caption{Example Annotations from PhyCheck}
\label{fig0}
\end{figure}

Building World Models is treated as a path towards achieving artificial general intelligence (AGI)~\cite{Lecun22}. World Models are AI systems that understand and reason about the physical working mechanisms of the real world~\cite{ding2025understanding}. Video modality has naturally become the primary testbed for exploring World Models, since this is the unique modality encompassing spatiotemporal dynamics reflecting physical principles, catalyzing the rapid proliferation of video understanding models, particularly the recent emergence of highly capable Video Large Language Models (Video-LLMs).

Existing Video-LLMs have exhibited strong commonsense reasoning and planning capabilities~\cite{physbench, yue2024mmmu, lu2024mathvista, kim2024openvlaopensourcevisionlanguageactionmodel, niu2024llarvavisionactioninstructiontuning, zhen20243dvla3dvisionlanguageactiongenerative}, but they have been found lacking the understanding of the physical world~\cite{physbench}. It is widely argued that current Video-LLMs fundamentally act as a sophisticated pattern recognizers driven by spurious statistical correlations, rather than mastering the causal relationships of underlying physical laws~\cite{motamed2026generative}--These models learn shortcuts without understanding. 

This deficiency in physical understanding not only impedes the transition of these models to World Models capable of internalizing and simulating reality, but also limits their deployment in Embodied AI scenarios where safe and reliable interactions are critical~\cite{physbench, wang2023newton, liu2024mokaopenworldroboticmanipulation, guo2024phygraspgeneralizingroboticgrasping}. For embodied AI to operate safely and reliably, and for current video understanding models to evolve into world models, mastering underlying physical laws (such as conservation, optical and mechanics laws) serves as an indispensable cornerstone for current video models.

Although recent studies have begun to recognize the importance of physical world understanding, the field still warrants deeper and more multifaceted exploration. Existing works primarily focus on evaluating current Video-LLMs' performance in terms of understanding physical world or interpreting physical commonsense. PhysGame proposed a benchmark to evaluate physical commonsense violations in gameplay videos~\cite{PhysGame}. PhysBench pointed out that VLMs' ability to comprehend physical phenomena remains limited~\cite{physbench}, and they benchmark numerous VLMs to show their deficiencies in understanding the physical world, likely due to the absence of physical knowledge in training data. They further incorporated vision foundation models and a physics knowledge memory. There exist works studying other perspectives of video understanding (e.g., MVPBench evaluates complex interaction capability across multiple videos~\cite{bai2026mvpbenchmultivideoperceptionevaluation}).

Although recent studies have begun to recognize the importance of physical world understanding, the field still warrants deeper and more multifaceted exploration. 
In this paper, we propose \textbf{PhyCheck}, a fine-grained, evidence-grounded video dataset designed for physical law understanding in Video Large Language Models (Video-LLMs). 
Unlike existing datasets that rely on coarse-grained overall descriptions or isolated judgments of physical common sense, PhyCheck systematically decomposes complex physical laws into explicit visual evidence clues—such as identifying spontaneous object movement or unexpected physical deformation—backed by rigorous human annotations (see Figure~\ref{fig0} for an example). 
To ensure maximum reliability, we constructed a large-scale collection of approximately 50k human-verified VQA pairs. Detailed descriptions of our data construction, including video collection, annotation expansion, and rigorous human verification protocol, are provided in PhyCheck Dataset section. 

In particular, PhyCheck bridges key gaps in current physical-understanding benchmarks through three distinctive design principles:

\textbf{(1) Counter-Physical Anomaly Focus:} 
Conventional datasets often rely on real-world videos depicting standard physical scenarios. The absence of counter-physical data restricts systematic evaluation and improvement of a model’s ability to differentiate between physically consistent and inconsistent dynamics. When a dataset contains only physically plausible videos, models may attain strong performance by exploiting statistical biases and superficial correlations in the data, rather than learning the underlying physical principles that govern real-world interactions. As a result, apparent performance gains may reflect shortcut learning or memorization instead of genuine physical understanding. PhyCheck emphasizes explicit physical law violations. By strategically combining synthetic anomaly videos—built upon VideoPhy2~\cite{videophy2}—with curated real-world clips, our dataset provides a balanced set of positive and negative samples. This setup not only rigorously evaluates Video-LLMs on counter-physical edge cases, but also offers contrastive perspectives to effectively fine-tune and align model capabilities.

\textbf{(2) Multi-Question Evidence-Grounded Decomposition:} 
Instead of pairing each video with an isolated, high-level query (e.g., asking ``\textit{which car has a higher average speed?}'' given a video of moving toy cars), PhyCheck equips every video sequence with a structured suite of multi-grained questions. We assume coarse-grained annotations are insufficient for capturing fine-grained physical phenomena and rule-specific inconsistencies. This lack of granularity constrains the development and assessment of models’ physical reasoning capabilities. To address these limitations, we construct a hierarchical VQA-based annotation system. There is a major question querying the underlying physical principle related to the video content, followed by a suite of fine-grained questions probing explicit visual evidence. These question-and-answer pairs annotates both physically consistent and physically violating events at the level of specific physical principles—such as spontaneous motion, unnatural acceleration, or unexpected structural deformation- which forms an evidence-grounded reasoning chain before reaching a physical verdict. This multi-question evidence-grounded annotation structure enables more precise supervision and more rigorous evaluation of Video-LLMs' physical understanding capabilities.

\textbf{(3) Context-Sensitive Physical Evaluation:} 
During the construction of PhyCheck, we realized that physical validity is often context-dependent. For instance, a floating ball appears physically impossible in isolation, yet becomes entirely plausible when contextual cues—such as an active hairdryer underneath or a zero-gravity environment—are present. PhyCheck incorporates context-aware video pairs to uniquely assess whether Video-LLMs can dynamically adjust their physical logic based on environmental cues.

Based on PhyCheck, we systematically evaluate numerous Video-LLMs covering both open-source and commercial models across different sizes. Our experiments reveal that existing video-LLMs generally exhibit superficial physical understanding when encountering physically violating videos. Furthermore, we propose a reasoning-chain-based fine-tuning approach to enhance the physical comprehension capability of Qwen2.5-VL~\cite{bai2025qwen25vltechnicalreport}. 

Our main contributions are summarized as follows:
\begin{enumerate}

\item We introduce \textbf{PhyCheck}, a fine-grained, evidence-grounded video dataset specifically engineered for physical law understanding. To the best of our knowledge, \textbf{PhyCheck} is the first video QA dataset that explicitly probes both compliance with and violation of physical laws, forcing Video-LLMs to transcend superficial visual priors and evaluate true physical causality.

\item We design a multi-tier VQA annotation system that serves a dual purpose: benchmarking and model alignment. Each instance pairs a video clip with a high-level primary question evaluating physical law compliance, alongside a suite of fine-grained questions on explicit visual evidence. While the primary questions rigorously diagnose model capabilities, the fine-grained evidence chains serve as structured reasoning scaffolding that enables supervised fine-tuning for enhanced physical comprehension.

\item We construct a specialized context-sensitive subset to evaluate whether Video-LLMs can dynamically update their physical law judgments upon integrating environmental context. This suite goes beyond direct, static consistency recognition to assess higher-level contextual physics reasoning.

\item Through extensive evaluation across a diverse spectrum of open-source, proprietary, and specialized Video-LLMs, we reveal severe systemic limitations in current models' physical intuition. 
Specifically, while current Video-LLMs demonstrate strong performance on standard positive samples, their performance drops sharply on negative violation samples—highlighting an acute vulnerability to counter-physical anomalies.
\end{enumerate}
\section{Related Work}

\subsection{Video LLMs}

Video LLMs extend vision-language models to video inputs by integrating temporal visual information with language reasoning. Recent studies have advanced video LLMs from multiple perspectives. Some works focus on enhancing video understanding capabilities: General-purpose video LLMs, Qwen and Gemini, provide broad multimodal instruction following and enhance complex reasoning over long and diverse video contexts. Other studies explore improved video modeling strategies, ranging from the compact spatiotemporal representations of LLaVA-OneVision2~\cite{llavaonevision2}, which better preserve motion cues, to unified frameworks such as UniVid~\cite{univid} and UniVideo~\cite{univideo}, which integrate video understanding and generation within a shared architecture. Despite these advances, existing video LLMs mainly target semantic and temporal comprehension, whereas physical understanding requires models to go beyond observed events to infer hidden mechanisms and assess their consistency with physical laws.

\subsection{Benchmarks for Evaluating Physical Understanding in Video-LLMs} 
To investigate whether Video LLMs can capture physical knowledge, several benchmarks have been proposed to evaluate physical reasoning capabilities from different perspectives. PhysBench~\cite{physbench} provides a comprehensive evaluation of Vision-Language Models' understanding of the physical world, covering object properties, relationships, scene-level reasoning, and physics-based dynamics. However, it mainly focuses on general physical knowledge rather than identifying violations of physical laws in dynamic scenarios. PhysGame~\cite{PhysGame} investigates whether video-language models can recognize physical commonsense violations in gameplay videos, such as abnormal motion and inconsistent interactions, but its evaluation is limited to game environments and lacks analysis of the underlying physical principles. MVP~\cite{MVP} proposes a shortcut-aware evaluation protocol based on minimal video pairs, aiming to distinguish genuine physical reasoning from reliance on superficial visual correlations. In short, existing benchmarks primarily provide coarse-grained evaluations of physical consistency, while fine-grained understanding of which physical laws are violated and what visual evidence supports such judgments remains underexplored.

\subsection{Video Generation Models and Physical Consistency Evaluation}

Recent breakthroughs in generative architectures have driven the rapid evolution of video generation models (e.g., Sora~\cite{videoworldsimulators2024}, Cosmos~\cite{nvidia2025cosmosworldfoundationmodel}, and Wan~\cite{wan2025wanopenadvancedlargescale}), enabling the synthesis of highly realistic and temporally coherent dynamic scenes.

As video synthesis advances toward simulating real-world dynamics, evaluating whether generated videos respect physical reality has become a critical research topic. Benchmarks such as VideoPhy~\cite{videophy} and VideoPhy2~\cite{videophy2} investigate whether video generation models can produce physically plausible events across diverse real-world scenarios. Concurrently, evaluation frameworks like VBench2~\cite{vbench2} and PhyGenBench~\cite{phygenbench} systematically incorporate physics-related metrics to benchmark video generation quality.

While these efforts focus on the \textit{synthesis side} by assessing whether generative models can render visually and physically plausible outcomes, our work targets a fundamentally distinct and complementary objective on the \textit{comprehension side}. 
Rather than evaluating the perceptual realism of synthesized frames, \textbf{PhyCheck} investigates whether Video-LLMs can acquire an intrinsic understanding of the underlying physical mechanisms and explicitly reason about the physical laws governing both valid and anomalous visual outcomes.
\section{PhyCheck Dataset}~\label{sec:dataset}

\subsection{Overview}

PhyCheck is a physics-laws-oriented video question answering dataset designed to evaluate and improve the ability of video understanding models to judge \textbf{whether observed phenomena comply with special physical laws}. It comprises \textbf{6,399} synthetic videos and contains \textbf{69,825} question–answer pairs, with an average of approximately \textbf{10.9} questions per video. Each question–answer pair is associated with one of six major categories of physical principles: (1) Mechanics Laws, (2) Conservation Laws, (3) Material Properties, (4) Fluids and Interface Phenomena, (5) Optical Laws, and (6) Thermal Laws‌. Notably, compared to the existing PhysGame and PhysBench dataset, all questions are presented in a binary yes/no format rather than as four-way multiple-choice questions, requiring models to directly determine whether a specific physical description is consistent with the observed video content.

\begin{table}[t]
\centering
\label{tab:physics_categories}
\renewcommand{\arraystretch}{1}
\begin{tabular}{p{0.34\linewidth} p{0.57\linewidth}}
\toprule
\textbf{Category} & \textbf{Representative laws} \\
\midrule
Mechanics Laws
& Gravity, friction, inertia, and Newton's first/second/third law \\

Conservation Laws 
& Conservation of momentum, conservation of mass, conservation of angular momentum, and conservation of energy \\

Material Properties
& Elasticity and hardness \\

Fluids and Interface Phenomena 
& Buoyancy, surface tension, and fluid dynamics \\

Optical Laws
& Reflection \\

Thermal Laws‌ 
& Melting and flame reactions \\

\bottomrule
\end{tabular}
\caption{Physical categories and representative laws.}
\end{table}

\subsection{Dataset Construction Framework}
\begin{figure*}[ht]
\centering
\includegraphics[width=\linewidth]{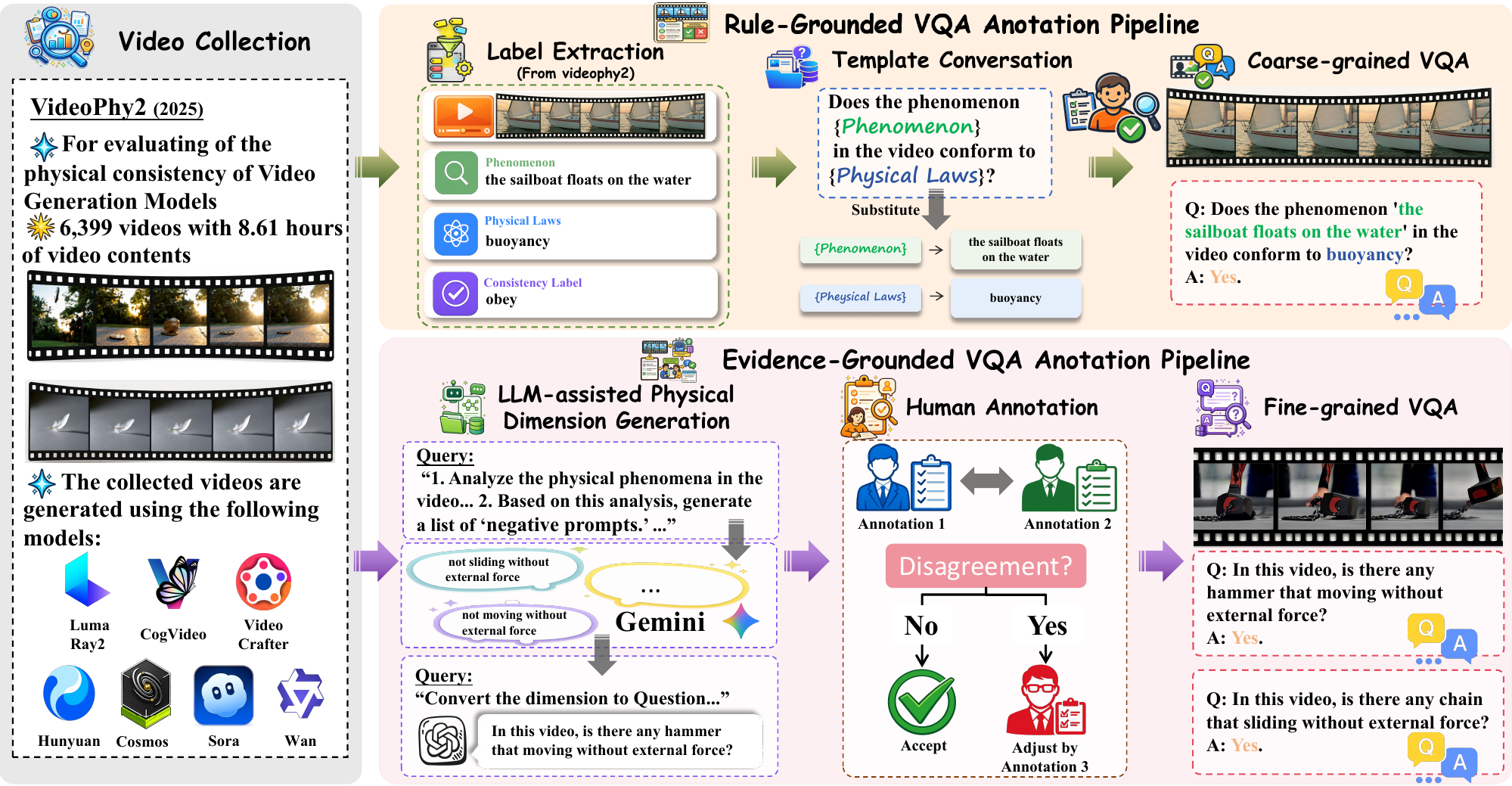}
\caption{Overview of the PhyCheck dataset construction framework.}
\label{fig1}
\end{figure*}
\paragraph{Data Source.} 
We collect synthetic video samples from VideoPhy2, a large-scale benchmark for evaluating physical consistency in generated videos. Specifically, we reuse the videos from the training and test splits of VideoPhy2, resulting in a total of 6,399 videos with 8.61 hours of video content. These videos are generated by seven representative text-to-video generation models, including (1) Luma Ray2~\cite{ray2}, (2) CogVideo~\cite{yang2025cogvideox}, (3) Cosmos~\cite{nvidia2025cosmosworldfoundationmodel}, (4) Hunyuan~\cite{kong2025hunyuanvideosystematicframeworklarge}, (5) VideoCrafter~\cite{chen2024videocrafter2}, (6) Sora~\cite{videoworldsimulators2024}, and (7) Wan~\cite{wan2025wanopenadvancedlargescale}. 

While VideoPhy2 released these raw video files originally created for benchmarking video generation models, we substantially extend their utility for physical comprehension assessment and improvement. Instead of relying on their high-level generation-side video summary descriptions, we re-annotate and restructure these videos into a fine-grained, evidence-grounded VQA dataset through rigorous multi-tier prompt engineering and 556 working hours of intensive human verification.

\paragraph{Rule-Grounded Physical VQA Construction Pipeline.} 
To construct VQA pairs that determine whether a video phenomenon conforms to or violates physical rules, we develop a rule-grounded physical VQA construction pipeline, as shown in Figure~\ref{fig1}.We first leverage the physical rule annotations provided by VideoPhy2, where each synthetic video is associated with candidate physical rules and corresponding labels indicating whether the observed phenomenon is followed, violated, or undetermined. Since ambiguous cases provide unreliable supervision, we retain only the followed and violated samples and discard the undetermined ones.

Rather than directly transforming the original rule descriptions into questions, we adopt a template-based generation strategy to ensure consistent question formulation and balanced supervision. Specifically, we design two types of rule-aware templates: (1) consistency templates, which ask whether the observed phenomenon conforms to a given physical rule, and (2) violation templates, which query whether the phenomenon violates the corresponding rule. The answers are deterministically assigned according to the original VideoPhy2 annotations. For consistency templates, followed cases are assigned ``Yes'' while violated cases are assigned ``No''. For violation templates, the answer mapping is reversed. This design balances question polarity and reduces potential biases caused by fixed answer patterns, while preserving the explicit correspondence between each question and its underlying physical law.

Each sample consists of a video, a rule-oriented question, and a binary answer, with the associated physical law retained as metadata for fine-grained analysis. When a video contains multiple valid physical rule annotations, we generate multiple VQA instances, each corresponding to a specific phenomenon--rule pair. Finally, all questions are manually reviewed to ensure that they are visually grounded, unambiguous, and answerable based on the provided video content.

\paragraph{Evidence-Grounded VQA Annotation Pipeline.}
Our fine-grained annotation pipeline aims to identify interpretable physical details that contribute to the violation or compliance of physical laws, thereby providing more informative supervision for improving VLMs' physical reasoning ability. Different from coarse-grained physical consistency labels that only indicate whether a phenomenon follows or violates a physical rule, fine-grained annotations focus on the specific visual and physical cues underlying the decision. The annotation process combines LLM-assisted physical reasoning with human verification, where large multimodal models generate candidate physical details, followed by manual annotation to ensure reliability.

Specifically, we first employ Gemini-3 to analyze each video and generate detailed descriptions of the observed physical phenomena. Given the video description and the associated physical law labels from VideoPhy2, Gemini-3 is further instructed to act as an expert in video generation and analyze potential physical inconsistencies. Based on this analysis, the model produces negative prompts that describe specific physical details which should be avoided when generating physically correct videos. The generated negative prompts which  serve as interpretable hypotheses of possible violations are then provided to GPT-5 for refinement. GPT-5 transforms each physical inconsistency description into a clear and answerable question format, asking whether a specific physical detail exists in the video. This process converts implicit physical reasoning cues into structured fine-grained VQA annotations while maintaining their connection to the corresponding physical laws.

Finally, all generated questions and candidate physical details undergo human verification. Each video-question pair is independently reviewed by two annotators to determine whether the queried detail is supported by the video. Samples with agreement are directly labeled, while disagreements are resolved by a third annotator through majority voting. Annotators also check whether the question accurately captures the underlying physical phenomenon and provides meaningful evidence for physical consistency or violation. To evaluate the reliability of human verification, we measure the agreement between the two initial annotators. They achieve a raw agreement rate of 99.74\% and a Cohen's $\kappa$ coefficient of 0.9959, indicating almost perfect inter-annotator agreement.  The resulting dataset contains video-level binary judgments together with fine-grained physical detail annotations, enabling both physical consistency evaluation and detailed analysis of Video-LLM behaviors.
\subsection{Context-Sensitive Physical Evaluation (pilot study)}

\begin{figure}[ht]
\centering
\includegraphics[width=\linewidth, height=8cm]{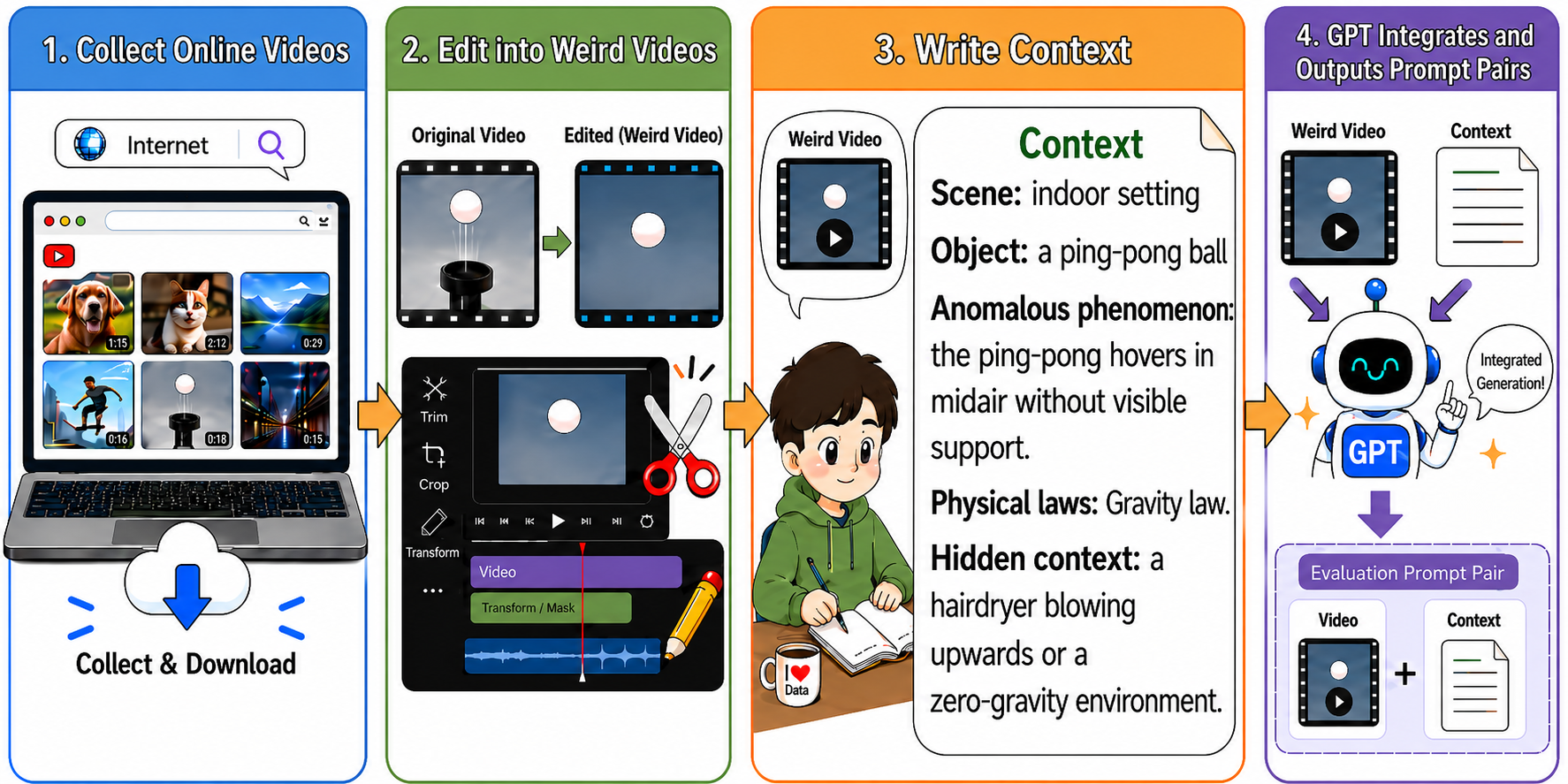} 
\caption{Overview of the context-sensitive dataset construction framework.}
\label{fig2}
\end{figure}

Beyond universal physical compliance, real-world physical validity can be context-dependent due to partial observations. 
To probe whether Video-LLMs can dynamically update their judgments given external causal context, we curate a small-scale pilot subset of 50 hand-crafted pairs (kept compact due to intensive manual editing, as primary efforts were dedicated to the main PhyCheck dataset). 
Specifically, we select real-world plausible videos and crop out the visible causal factors (e.g., removing a hairdryer to leave a floating ball), rendering the cropped clip seemingly implausible. 
We then supplement each clip with a manually written context describing the cropped part, namely the hidden cause (e.g., upward airflow). 
Models are evaluated on whether they can recognize the implausibility of the isolated visual effect alone and update their judgments once provided with external causal context (see Figure~\ref{fig2}).

\section{Model Evaluation and Enhancement}

\begin{table*}[t]
\centering
\setlength{\tabcolsep}{4pt}
\renewcommand{\arraystretch}{1}
\begin{tabular}{l|c|c ccc ccc}
\toprule
& & Overall & \multicolumn{3}{c}{Conform Example}
& \multicolumn{3}{c}{Voilate Example} \\
\cmidrule(lr){3-3}
\cmidrule(lr){4-6}
\cmidrule(lr){7-9}
Model
& Params
& Accuracy $\uparrow$
& Precision $\uparrow$ & Recall $\uparrow$ & F1 $\uparrow$
& Precision $\uparrow$ & Recall $\uparrow$ & F1 $\uparrow$ \\
\midrule

\begin{tabular}[c]{@{}c@{}}
  Qwen3-Instruct
  \end{tabular}
  & 8B
& 62.82 & 67.41 & 84.68 & 75.06 & 40.41 & 20.24 & 26.97 \\

  \begin{tabular}[c]{@{}c@{}}
  Qwen3-Thinking
  \end{tabular}
  & 8B
& 60.28 & 66.64 & 79.91 & 72.67 & 36.03 & 22.05 & 27.36 \\

\begin{tabular}[c]{@{}c@{}}
Qwen3.5
\end{tabular}
& 9B
    & 59.42
    & 65.91
    & 79.92
    & 72.24
    & 33.23
    & 19.46
    & 24.55 \\

\begin{tabular}[c]{@{}c@{}}
Qwen3.5
\end{tabular}
& 27B
    & 58.81
    & 65.71
    & 78.80
    & 71.66
    & 32.48
    & 19.87
    & 24.66 \\

\begin{tabular}[c]{@{}c@{}}
Qwen3.6
\end{tabular}
& 27B
    & 57.77
    & 65.65
    & 75.70
    & 70.32
    & 32.53
    & 22.83
    & 26.83 \\

\begin{tabular}[c]{@{}c@{}}
  LlaVa-OneVision2
  \end{tabular}
  & 8B
& \textbf{63.11} & 66.68 & \textbf{88.28} & \textbf{75.98} & 38.12 & 14.07 & 20.55 \\

  \begin{tabular}[c]{@{}c@{}}
  UniVid
  \end{tabular}
  & 7B
& 59.79 & 66.18 & 80.05 & 72.46 & 34.32 & 20.31 & 25.51 \\

\begin{tabular}[c]{@{}c@{}}
UniVideo
\end{tabular}
& 7B
& 50.93 & 65.73 & 53.78 & 59.16 & 33.51 & \textbf{45.38} & \textbf{38.55} \\

\begin{tabular}[c]{@{}c@{}}
Gemini-3.5-Flash
\end{tabular}
& -
& 62.24
    & 67.05
    & 84.29
    & 74.69
    & 38.65
    & 19.28
    & 25.73 \\

\begin{tabular}[c]{@{}c@{}}
Skyra-SFT
\end{tabular}
& 7B
    & 61.57
    & \textbf{68.47}
    & 77.59
    & 72.74
    & \textbf{41.03}
    & 30.38
    & 34.91 \\

\begin{tabular}[c]{@{}c@{}}
Skyra-RL
\end{tabular}
& 7B
    & 61.31
    & 68.38
    & 77.11
    & 72.48
    & 40.64
    & 30.54
    & 34.87 \\

\bottomrule
\end{tabular}

\caption{Overall Physical Understanding Evaluation results of current Video-LLMs on PhyCheck.}

\label{tab:model_results}
\end{table*}

\begin{table*}[t]
\centering
\setlength{\tabcolsep}{3pt}
\renewcommand{\arraystretch}{1}

\begin{tabular}{cc|ccccccc}
\toprule
Coarse-grained& Fine-grained & Overall & \multicolumn{3}{c}{Conform Example}
& \multicolumn{3}{c}{Violate Example} \\
\cmidrule(lr){3-3}
\cmidrule(lr){4-6}
\cmidrule(lr){7-9}
VQA & VQA
& Accuracy $\uparrow$
& Precision $\uparrow$ & Recall $\uparrow$ & F1 $\uparrow$
& Precision $\uparrow$ & Recall $\uparrow$ & F1 $\uparrow$ \\
\midrule

&
& 50.93 & 65.73 & 53.78 & 59.16 & 33.51 & 45.38 & 38.55 \\

& $\checkmark$
      & 51.43
      & 65.49
      & 56.00
      & 60.38
      & 33.15
      & 42.51
      & 37.26 \\

$\checkmark$ & 
      & 66.83
      & 77.02
      & 70.99
      & 73.88
      & 50.96
      & \textbf{58.73}
      & 54.57 \\

\midrule
$\checkmark$ $\textcircled{1}$ & $\checkmark$ $\textcircled{2}$
      & 51.89
      & 66.67
      & 54.39
      & 59.91
      & 34.61
      & 47.04
      & 39.88 \\
$\checkmark$ $\textcircled{2}$ & $\checkmark$ $\textcircled{1}$
 & \textbf{81.00}
      & \textbf{80.68}
      & \textbf{93.68}
      & \textbf{86.69}
      & \textbf{82.05}
      & 56.30
      & \textbf{66.78} \\
\bottomrule
\end{tabular}
\caption{Dataset Utility Validation of PhyCheck with different supervision settings.}
\label{tab:finetune_results}
\end{table*}
\subsection{Settings}
\paragraph{Evaluated Models.} 
Our evaluation covers several representative open-source and proprietary Video-LLMs, comprising the Qwen series, Gemini-3.5-Flash, LLaVA-OneVision2, UniVid, and UniVideo, spanning diverse video understanding paradigms. UniVid and UniVideo are unified video models that reuse existing video understanding backbones, based on BAGEL-7B-MoT~\cite{deng2025emergingpropertiesunifiedmultimodal} and Qwen2.5-VL, respectively. Additionally, Skyra~\cite{skyra} is specifically designed for AI-generated video detection through grounded artifact reasoning. It identifies and localizes physics-related inconsistencies caused by violations of real-world physical principles, providing a specialized baseline for physical consistency understanding.
\paragraph{Evaluation Setup.} 
\textbf{(1) Direct Evaluations.} We first directly evaluate existing Video-LLMs on the PhyCheck test dataset to investigate whether current models can verify physics-related statements grounded in video content. \textbf{(2) Supervised Fine-Tuning Utility.} To validate the training utility of PhyCheck, we fine-tune Qwen2.5-VL on the training split and compare its performance with existing baselines under the same evaluation protocol. \textit{Exploratory Pilot Probe:} 
In addition, we conduct a lightweight pilot probe on our 50-pair context-sensitive subset. Using a three-setting evaluation protocol (video-only, context-assisted, and joint accuracy), we evaluate whether models can dynamically adjust their physical judgments upon integrating external causal context, rather than relying on rigid visual shortcuts. 

\paragraph{Metrics.} For the main evaluation on PhyCheck, We report overall accuracy, class-wise precision, recall, and F1 scores for both Conform and Violate examples. These metrics provide a comprehensive evaluation of models' ability to verify physical statements and reveal potential prediction biases toward either category. 
\paragraph{Implementation Details.} We adopt Qwen2.5-VL as the base model for its open-source availability and broad adoption in video understanding studies. It is fully fine-tuned on the PhyCheck training set for 12 epochs. All experiments are conducted using PyTorch on NVIDIA A100 GPUs.

\begin{figure*}[ht]
\centering
\includegraphics[width=\linewidth]{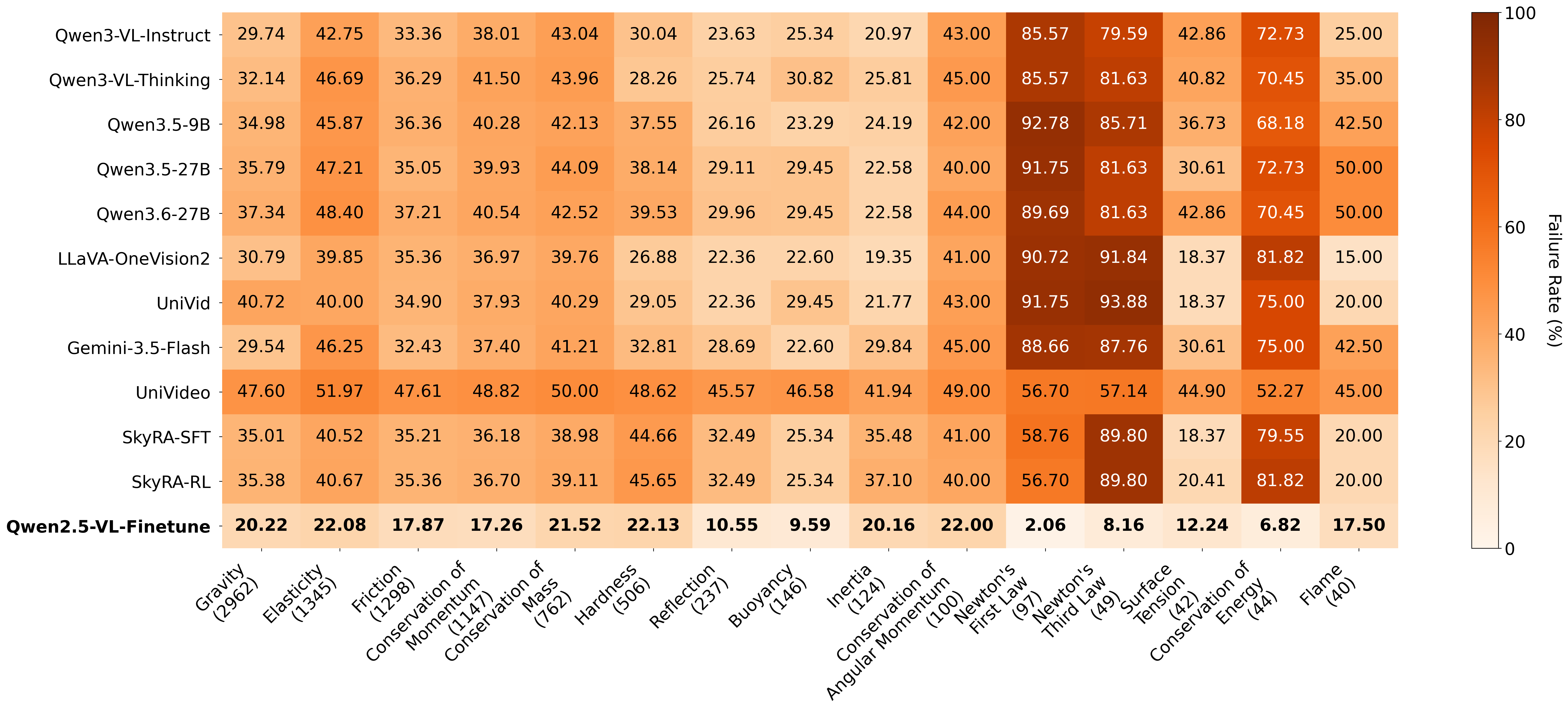}
\caption{
Failure Rates across Physical Law Categories (Top 15). Parentheses indicate sample counts.
}
\label{fig3}
\end{figure*}

\subsection{Overall Physical Understanding of Video-LLMs}
As shown in Table~\ref{tab:model_results}, existing models achieve moderate performance on PhyCheck, with most models obtaining accuracies around 60\%, indicating that current Video-LLMs struggle with reliable physical-law judgment. Beyond overall accuracy, we analyze class-specific metrics for Conform and Violate examples. Most models exhibit higher recall on Conform examples than Violate examples, indicating a conservative prediction tendency when verifying physics-related statements. Skyra achieves competitive performance despite being primarily designed for AI-generated video detection through artifact reasoning. However, its performance does not consistently surpass general-purpose Video-LLMs, indicating that detecting physics-related artifacts is not equivalent to understanding underlying physical laws.

\subsection{Dataset Utility Validation}
To investigate the contribution of different supervision components in PhyCheck, we conduct a dataset utility analysis by fine-tuning Qwen2.5-VL with different subsets and training strategies, including coarse-grained VQA, fine-grained VQA, and their sequential combinations. As shown in Table ~\ref{tab:finetune_results}, different supervision settings lead to distinct performance variations.

Training with fine-grained VQA alone provides only marginal improvement over the original model, with accuracy increasing from 50.93\% to 51.43\%. This suggests that fine-grained questions alone may not provide sufficient supervision for establishing fundamental physical consistency understanding. In contrast, training with coarse-grained VQA achieves a substantial improvement, increasing accuracy to 66.83\%, highlighting the effectiveness of coarse-grained physical consistency supervision in providing foundational signals for physical understanding. Furthermore, we investigate the impact of combining different supervision components through sequential training. Training with coarse-grained VQA followed by fine-grained VQA results in limited improvement, whereas reversing the training order achieves the best performance of 81.00\%. These results indicate that learning fine-grained physical consistency before coarse-grained physical reasoning provides a more effective learning trajectory, where fine-grained supervision establishes the learned physical concepts and coarse-grained supervision further enhances physical reasoning capability. Overall, these results validate the effectiveness of the hierarchical supervision design in PhyCheck, demonstrating that different annotation granularities provide complementary signals for improving Video-LLMs' physical law understanding.

\subsection{Category-wise Failure Analysis}
\begin{table}[t]
\centering
\scriptsize
\setlength{\tabcolsep}{12pt}
\renewcommand{\arraystretch}{1}

\begin{tabular}{l|ccccccc}
\toprule
\multicolumn{1}{c}{\textbf{Model}}
& \multicolumn{1}{c}{
    \begin{tabular}[c]{@{}c@{}}
    \textbf{VideoCrafter}\\
    \hline
    \textbf{1984}
    \end{tabular}
}
& \multicolumn{1}{c}{
    \begin{tabular}[c]{@{}c@{}}
    \textbf{Wan}\\
    \hline
    \textbf{1893}
    \end{tabular}
}
& \multicolumn{1}{c}{
    \begin{tabular}[c]{@{}c@{}}
    \textbf{Cosmos}\\
    \hline
    \textbf{1889}
    \end{tabular}
}
& \multicolumn{1}{c}{
    \begin{tabular}[c]{@{}c@{}}
    \textbf{CogVideo}\\
    \hline
    \textbf{1866}
    \end{tabular}
}
& \multicolumn{1}{c}{
    \begin{tabular}[c]{@{}c@{}}
    \textbf{Ray2}\\
    \hline
    \textbf{1237}
    \end{tabular}
}
& \multicolumn{1}{c}{
    \begin{tabular}[c]{@{}c@{}}
    \textbf{Hunyuan}\\
    \hline
    \textbf{403}
    \end{tabular}
}
& \multicolumn{1}{c}{
    \begin{tabular}[c]{@{}c@{}}
    \textbf{Sora}\\
    \hline
    \textbf{180}
    \end{tabular}
}\\
\midrule

\begin{tabular}[c]{@{}c@{}}
Qwen3-Instruct
\end{tabular}
& 43.40
& 31.59
& 37.37
& 33.39
& 36.22
& 52.36
& 37.22 \\

\begin{tabular}[c]{@{}c@{}}
Qwen3-Thinking
\end{tabular}
& 45.26
& 36.29
& 39.07
& 35.42
& 39.37
& 51.36
& 42.22 \\

\begin{tabular}[c]{@{}c@{}}
Qwen3.5-9B
\end{tabular}
& 45.46
& 35.71
& 40.18
& 37.30
& 40.99
& 55.83
& 39.44 \\

\begin{tabular}[c]{@{}c@{}}
Qwen3.5-27B
\end{tabular}
& 46.77
& 35.97
& 40.50
& 38.00
& 42.12
& 54.84
& 37.78 \\

\begin{tabular}[c]{@{}c@{}}
Qwen3.6-27B
\end{tabular}
& 46.52
& 38.14
& 41.34
& 39.92
& 42.60
& 55.33
& 39.44 \\

\begin{tabular}[c]{@{}c@{}}
LlaVa-OneVision2
\end{tabular}
& 42.99
& 32.07
& 36.42
& 31.73
& 37.11
& 56.08
& 34.44 \\

\begin{tabular}[c]{@{}c@{}}
UniVid
\end{tabular}
& 45.51
& 36.61
& 39.44
& 36.07
& 39.69
& 53.10
& 45.56 \\

\begin{tabular}[c]{@{}c@{}}
UniVideo
\end{tabular}
&48.84
&49.02
&49.07
&49.20
&48.75
&51.36
&47.78 \\

\begin{tabular}[c]{@{}c@{}}
Gemini-3.5-Flash
\end{tabular}
& 43.20
& 32.59
& 38.06
& 34.03
& 37.35
& 54.09
& 34.44  \\

\begin{tabular}[c]{@{}c@{}}
Skyra-SFT
\end{tabular}
& 41.99
& 35.97
& 39.92
& 33.28
& 39.13
& 45.66
& 41.67  \\

\begin{tabular}[c]{@{}c@{}}
Skyra-RL
\end{tabular}
& 42.04
& 36.24
& 40.02
& 33.76
& 39.45
& 46.15
& 42.78  \\
\midrule
\begin{tabular}[c]{@{}c@{}}
Qwen2.5-Finetune
\end{tabular}
&\textbf{23.74}
&\textbf{16.22}
&\textbf{18.95}
&\textbf{17.42}
&\textbf{19.32}
&\textbf{15.14}
&\textbf{19.44} \\

\bottomrule
\end{tabular}
\caption{
Failure Rates by Generation Category. Subscripts denote evaluated video counts.
}
\label{tab:generation_source_results}
\end{table}

\paragraph{Physical Law Categories:} As shown in Figure~\ref{fig3}, the prediction failure rates vary substantially across different physical phenomena, suggesting that current Video-LLMs exhibit uneven capabilities across physical concepts rather than a unified understanding of physical laws. Off-the-shelf Video-LLMs generally achieve lower failure rates on visually intuitive phenomena, such as Inertia, Reflection, and Buoyancy, where relevant physical cues are directly observable from object motions and interactions. In contrast, higher failure rates are observed in categories requiring more implicit physical reasoning, including Conservation laws and Newtonian mechanics. These categories require models to infer latent physical states and reason about abstract relationships that cannot be directly captured from visual appearances alone. After fine-tuning on PhyCheck, Qwen2.5-VL substantially reduces failure rates across all physical categories, with particularly notable improvements on Conservation laws and Newtonian mechanics.
\paragraph{Video Generation Sources:} As shown in Table~\ref{tab:generation_source_results}, Video-LLMs show relatively consistent failure patterns across different video generation sources, indicating that the generation model is not the primary factor affecting physical law understanding. These results suggest that current failures mainly originate from insufficient physical reasoning capabilities rather than source-specific visual characteristics.

\subsection{Context-assisted physical Understanding.}
As an initial exploratory probe, we evaluate the model behaviors on our 50-pair context-sensitive subset (refer to results in Appendix C3). Baselines achieve high accuracy under the video-only setting (e.g., $0.96$ for Qwen3.5-27B), many struggle to update their predictions after context is introduced. This trend suggests a potential over-reliance on rigid visual shortcuts, where models tend to prematurely treat partial visual anomalies as definitive physical violations.

In contrast, our fine-tuned \textbf{Qwen2.5-Finetune} exhibits a distinct shift. Under the video-only setting, it shows a lower performance ($0.18$), which may reflect a form of epistemic caution. Having been trained on diverse physical rules, the model appears less inclined to strictly penalize visual observations. Remarkably, when context is provided, its performance rises to \textbf{0.98}. 

We admit these preliminary results should be interpreted cautiously considering the limited scale of this subset, but they tentatively indicate that model alignment on PhyCheck can foster context-sensitive reasoning, encouraging models to dynamically integrate hidden physical factors rather than adhering solely to surface visual cues.

\section{Conclusion}
In this work, we introduce \textbf{PhyCheck}, a physics-law-oriented video question answering dataset designed to evaluate and improve the physical law understanding ability of Video-LLMs. PhyCheck provides both fine-grained physical-law questions and a context-sensitive subset for evaluating physical judgments. Through comprehensive evaluation of representative open-source, proprietary, and specialized Video-LLMs, we reveal substantial limitations in current models' physical law understanding. Furthermore, fine-tuning experiments with Qwen2.5-VL demonstrate that PhyCheck provides effective supervision for improving physical understanding, highlighting its value as both an evaluation benchmark and a training resource for developing more physically grounded video understanding models.

\bibliography{main}
\bibliographystyle{plainnat}



\end{document}